\pdfoutput=1

\documentclass[11pt]{article}

\usepackage{acl}

\usepackage{times}
\usepackage{latexsym}

\usepackage[T1]{fontenc}

\usepackage[utf8]{inputenc}

\usepackage{microtype}

\usepackage{inconsolata}

\usepackage{graphicx}
\usepackage[inkscapelatex=false]{svg}

\usepackage{qtree}
\usepackage{enumerate}
\usepackage{changepage}
\usepackage{amsmath}
\newcommand{\cprime}{\textsc{Prime}}
\newcommand{\ctarget}{\textsc{Target}}
\newcommand{\vprime}{\textit{Prime}}

\title{LLMs syntactically adapt their language use to their conversational partner}

\author{Florian Kandra, Vera Demberg, Alexander Koller\\
        Department of Language Science and Technology\\
        Saarland University, Saarbrücken, Germany\\
        \texttt{flka00003@uni-saarland.de}\\
        \texttt{\{vera, koller\}@coli.uni-saarland.de}
        }

\begin{document}
\maketitle
\begin{abstract}
It has been frequently observed that human speakers align their language use with each other during conversations.
In this paper, we study empirically whether large language models (LLMs) exhibit the same behavior of conversational adaptation.
We construct a corpus of conversations between LLMs and find that two LLM agents end up making more similar syntactic choices as conversations go on, confirming that modern LLMs adapt their language use to their conversational partners in at least a rudimentary way.
\end{abstract}

\section{Introduction}

It has been documented broadly that when humans talk to each other, they adapt their language use to their communication partners by coordinating their behavior and language. Humans \emph{align} not only their gestures, posture, and speech rate \citep{Holler2011mimicry,shockley2009coordinative,jungers2009speech}, but also their linguistic decisions at deeper levels, such as semantics and syntax \cite{BOCK1986355,garrod1987semanticcoord}. In other words, the distribution over syntactic structures of two human speakers becomes more similar as a conversation progresses.

In this paper, we investigate whether large language models (LLMs) adapt their syntactic choices to their conversational partners as well.
While it is well known that LLMs can be explicitly prompted towards embodying different ``personas'' and changing the style of the language they generate \cite{deshpande-etal-2023-toxicity, thillainathan2025finegrained}, it is unclear whether merely being present in a conversation with an interlocutor is sufficient to make LLMs adapt their language use to their interlocutor's.
The ability to adapt to the communication partner's language is associated with increased success in goal-oriented conversations \cite{reitter2014}, and it enables a dialogue system to meet a user's language use rather than requiring the user to adapt to the system \cite{schlangen2022normparticipationgroundslanguage}.
Language models will only serve as effective foundations for dialogue systems if they prove capable of implicitly adapting to a user's language.

To this end, we create a new dataset of conversations between LLMs in which both LLMs are prompted to initially exhibit different language use.
We then measure the dynamics of syntactic language adaptation over the course of the conversations, using a method adapted from the human-human analysis of \newcite{reitter2014}.
We find that GPT-4o \cite{openai2024gpt4ocard} and Llama-3-8B \cite{grattafiori2024llama} conversations show statistically significant adaptation when comparing syntactic repetitions within conversations against repetitions across conversations, replicating Reitter's findings for human conversations.
We further show this is a continuous process active throughout conversations and conclude by discussing whether these findings demonstrate ``human-like'' alignment in LLMs.

Our code and data are publicly available\footnote{\url{https://github.com/coli-saar/llm-language-accomodation}}.

\section{Background} \label{sec:background}
As we mentioned above, humans adapt their language use to their communication partners across various linguistic levels.
In this paper, our focus is on \emph{syntactic} adaptation: Do the distributions over the syntactic structures that two interlocutors produce become more similar over the course of a conversation?

In the psycholinguistics literature on human communication, two separate (but not exclusive) mechanisms have been proposed to explain the mutual adaptation of language use.
\citet{Rasenberg2020framework} contrast two theoretical views that explain the process through \emph{alignment} on different cognitive levels: on a conscious level, in which cooperative decisions establish a situational common ground \cite{brennan1996conceptual_pacts}, and a subconscious level, in which automatic \emph{priming} leads to aligned representational states \cite{Pickering_Garrod_2004}.
In psycholinguistics, priming refers to a process in which encountering a word or construction temporarily increases the activation of a cognitive representation, thereby increasing the probability for the word or construction to be reproduced.

In this paper, we study the conversational behavior of artificial, LLM-based agents. We will primarily focus on the level of outwardly observable changes to the language use and describe it with the theory-neutral word \emph{adaptation}. We will discuss in Section~\ref{sec:discussion} the extent to which concepts like alignment and priming can apply to LLMs.

\paragraph{Related Work.} 

\citet{cai-etal-2024-large} examined syntactic adaptation in LLMs,
with a focus on short-term priming. They showed that ChatGPT and Vicuna are more likely to complete a sentence with a double object or a prepositional object when primed with a sentence of the respective type. 
We extend this research to long conversations with natural sentences rather than carefully constructed one-sentence stimuli.

\section{Measuring human-human adaptation}
The phenomenon of long-term syntactic adaptation was first measured on corpora of human-human dialogues by \newcite{reitter2014}. 
The basic idea is to determine whether the usage frequency of a syntactic structure (specifically, a rule in a context-free grammar) in the first half of a conversation has a statistical impact on its frequency in the second half.

\begin{figure}
  \begin{tabular}{p{0.25\textwidth}p{0.2\textwidth}}
    \resizebox{0.33\textwidth}{!}{
      \Tree[.S 
      [.NP \it{we} ]
      [.VP 
        [.V \it{gave} ]
        [.NP 
          [.Det \it{the} ]
          [.N \it{policeman} ]
          ]
          [.NP 
          [.Det \it{a} ]
          [.N \it{toy} ]
        ]
      ]
    ]}&
    \begin{center}
      \resizebox{0.2\textwidth}{!}{
        \begin{tabular}{cl}
          R1.&$\mathrm{S} \rightarrow \mathrm{NP}\; \mathrm{VP}$\\
          R2.&$\mathrm{VP} \rightarrow \mathrm{V}\; \mathrm{NP}\; \mathrm{NP}$\\
          R3.&$\mathrm{NP} \rightarrow \mathrm{Det}\; \mathrm{N} $
        \end{tabular}
        }
      \end{center}
    \end{tabular}
  \caption{Phrase Structure Tree and Extracted Production Rules}
  \label{fig:tree}
\end{figure}


We follow Reitter in splitting each conversation in a dialogue corpus into two parts. We call the first 49\% of each conversation \cprime\ and the last 49\% of each conversation \ctarget ; the middle 2\% are discarded to ensure that we measure long-term adaptation as opposed to short-term priming. 
On corpora that are not already syntactically annotated, we parse 
each conversation with the Neural Berkeley Parser\footnote{We used the benepar\_en3\_large model of the \href{https://pypi.org/project/benepar/}{benepar} python package for parsing and \href{https://spacy.io/}{spacy}'s en\_core\_web\_md model for tokenization.} \cite{kitaev-klein-2018-constituency, kitaev-etal-2019-multilingual}, to obtain a set of context-free production rules for the \cprime\ and \ctarget\ section of each conversation, respectively (see Fig. \ref{fig:tree}).

\begin{figure}
  \centering
  \includegraphics[width=0.45\textwidth]{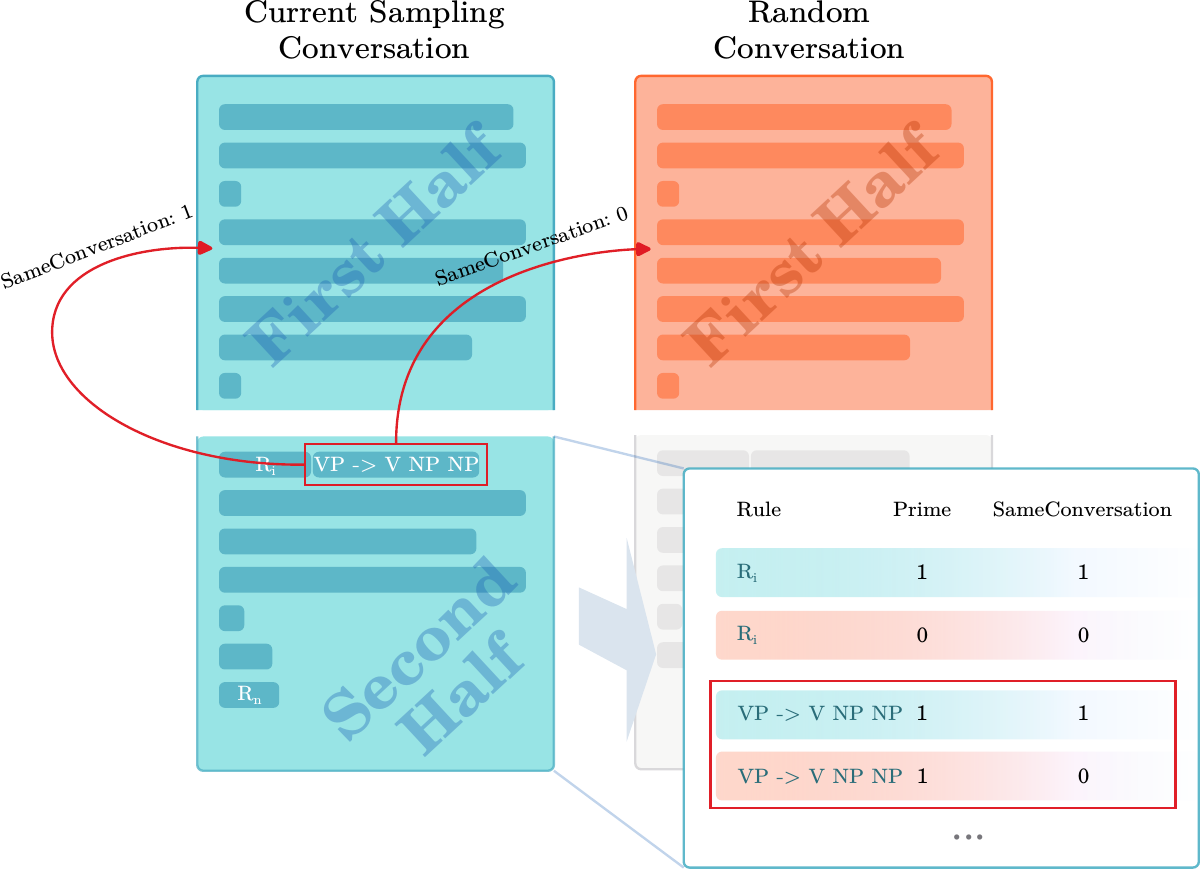}
  \caption{Sampling process to analyse syntactic alignment. Samples are drawn by checking rule occurences in the same conversation and in different random conversations.}
  \label{fig:sampling_process}
\end{figure}

Adaptation takes place if rule repetitions are more likely between the \cprime\ and \ctarget\ of the same conversation (where adaptation is possible), compared to a \cprime\ and \ctarget\ of different conversations (where no adaptation could have taken place).
In order to fit a model that can find such an effect, we sample data from our conversations in the following way:
For each rule across the \ctarget s of all conversations, we draw two samples, one for which we check whether that rule has been uttered within the same conversation, but by the other speaker, and one sample for which we check whether that rule has been uttered by a random speaker of a random other conversation.
Figure \ref{fig:sampling_process} depicts this process.
For every sample, we encode whether a rule was found this way in a binary variable \vprime.
Another binary variable, \textit{SameConv}, is used to indicate whether we looked for a prime in the same conversation ($1$) or in a different, random conversation ($0$).
If repetitions are more likely between speakers within conversations, such that we see an effect of \textit{SameConv} on \textit{Prime}, we take that as evidence of cross-speaker adaptation.

We further include features representing the log-frequency of rules across all conversations (\textit{ln(Freq)}), as more frequent rules are expected to be more likely to appear in any \cprime , and a variable \textit{ln(Size)}.
This second variable encodes the amount of different rules that a speaker used in the \cprime\ of a conversation, i.e.~the size of the set of rules that we use to look for a prime; a larger set increases the probability of any rule to occur.
We follow \citet{reitter2014} in excluding rules that appear only once in the whole dataset and rules that have disproportionately high frequencies (around $0.3\%$ of each dataset), because these rules are never primed or almost always primed. Including these rules in the analysis does not substantially change the results (see Appendix \ref{sec:analysis_all_rules}). We further remove structures that are lexically identical.

Our analysis differs from Reitter's original method in two aspects. First, we consider only overlaps between rule uses in \ctarget\ with uses in \cprime\ by the other speaker. This eliminates effects that solely stem from speaker idiosyncrasies or the conditioning of LLM-generated language on its own prior output. Second, our analysis includes the set size of rules used to check for a prime.

\subsection{Alignment in human conversations}\label{sec:reitter_human}
We replicate Reitter's results on human-human conversations to ensure that we obtain comparable results after our modifications.
We use the method described above to analyze the Switchboard corpus \cite{marcus1994penn}, which comprises 650 syntactically annotated telephone conversations (see Fig. \ref{fig:switchboard} in Appendix \ref{sec:datasets} for an overview of its composition).
This is in contrast to Reitter's work, which used the HCRC Map Task corpus \cite{anderson1991map}, consisting of task-oriented conversations.
By looking at Switchboard as opposed to Map Task, we demonstrate alignment effects on non-task-oriented conversations, facilitating comparison with LLM-generated conversations, and we make use of hand-annotated rather than automatically parsed syntactic structures.


We fit a mixed-effects logistic regression to the sampled data using the generalized linear mixed models (GLMM) of Python's \href{https://eshinjolly.com/pymer4/}{pymer4} (v0.8.2) package. We included a nested random intercept for conversations and speakers and a random slope for \textit{ln(Freq)} and centered fixed effects except \textit{SameConv}.
We selected the model through a backward selection process.
Results are shown in Table \ref{tab:analysis_results}.

\begin{table*}
  \centering
  \resizebox{0.96\textwidth}{!}{ 
      \begin{tabular}{ccc} 
      
        \begin{tabular}{lrrrr}
          &\multicolumn{4}{c}{Switchboard Corpus}\\
            \hline
            & $\beta$ & SE & $z$ & $p>|z|$ \\
            \hline
            Intercept & -2.927 & 0.018 & -158.8 & 0.000 \\
            ln(Freq) & 1.174 & 0.008 & 143.2 & 0.000 \\
            SameConv & 0.228 & 0.023 & 9.9 & 0.000 \\
            ln(Size) & 1.402 & 0.033 & 41.9 & 0.000 \\
            ln(Freq):SameConv & -0.101 & 0.01 & -9.8 & 0.000 \\
            ln(Freq):ln(Size) & 0.068 & 0.015 & 4.7 & 0.004 \\
            \hline
        \end{tabular}

        &
        \begin{tabular}{rrrr}
            \multicolumn{4}{c}{GPT Corpus}\\
            \hline
            $\beta$ & SE & $z$ & $p>|z|$ \\
            \hline
            -2.031 & 0.048 & -42.5 & 0.000 \\
            1.275 & 0.028 & 45.6 & 0.000 \\
            0.198 & 0.056 & 3.5 & 0.000 \\
            1.175 & 0.107 & 11.0 & 0.000 \\
            -0.146 & 0.035 & -4.2 & 0.000 \\
            0.266 & 0.062 & 4.3 & 0.000 \\
            \hline
        \end{tabular}

        &
        \begin{tabular}{rrrr}
          \multicolumn{4}{c}{Llama Corpus}\\
          \hline
          $\beta$ & SE & $z$ & $p>|z|$ \\
          \hline
          -2.016 & 0.049 & -41.0 & 0.000 \\
          1.333 & 0.030 & 44.5 & 0.000 \\
          0.505 & 0.052 & 9.6 & 0.000 \\
          0.825 & 0.147 & 5.6 & 0.000 \\
          -0.252 & 0.032 & -7.8 & 0.000 \\
          -0.005 & 0.088 & -0.06 & 0.952 \\
          \hline
        \end{tabular}
    \end{tabular}
  }
  \caption{\label{tab:analysis_results} The regression models for the Switchboard corpus (left) and the GPT corpus (middle) and LLama corpus (right). Effects show high significance except for the interaction between \textit{ln(Freq)} and \textit{ln(Size)} in the Llama corpus.}
\end{table*}

We find that \textit{SameConv} ($\beta=0.228, p<0.001$) has a significant positive effect, replicating Reitter's findings that humans align syntactically to their partners over the course of a conversation.


\section{Measuring LLM-LLM adaptation}\label{sec:experiment1}

We follow the same method to analyze syntactic adaptation in conversations of GPT-4o and Llama-3-8B.

\paragraph{Dataset.}
One challenge towards this goal is the availability of a suitable dataset of LLM conversations. We require a dataset consisting of long natural conversations (with no intervening task prompts) in which the speakers use varying syntactic structures to make adaptation possible and evenly distributed utterance lengths.

Existing datasets of conversations with LLMs do not satisfy these requirements.
UltraChat \cite{ding-etal-2023-enhancing} is a dataset of LLM-LLM conversations, but these conversations follow simple question-answering between a user and a model ``persona''. Conversations are too short and there is no variability between the language use across conversations.
By contrast, available datasets of human-LLM conversations, such as WildChat \cite{zhao2024wildchat1mchatgptinteraction}, consist of conversations that each have unique instructions by the user. This  makes conversations incomparable and therefore unsuitable for a statistical analysis of adaptation.

We therefore created our own dataset by letting GPT-4o and Llama-3-8B\footnote{We used GPT-4o-2024-08-06 and 
Meta-Llama-3-8B-Instruct with default parameters.} converse with themselves.
We created 17 different conversational agents with identical system prompts, except for an initial specification of a ``language persona'' that is unique to each agent.
We then generated conversations between pairs of LLM agents by iteratively prompting each of them for the next utterance, including the context of the conversation history.
Iterations were stopped, once a conversation surpassed a predefined length threshold.
All prompts for managing the conversations and defining the language personas can be found in Appendix \ref{sec:prompts}.

To ensure sufficient variety in the agents' language use, we further generated conversations where each agent conversed with itself.
We then calculated how often each syntactic rule was used and normalized these frequencies to create a discrete probability distribution of syntactic rules for each agent. 
To compare these distributions, we measured their distances using the Jensen-Shannon divergence (JSD). See Figures \ref{fig:gpt_jsdscores} and \ref{fig:llama_jsdscores} in Appendix \ref{sec:base_jsd} for details.
The results confirm a high degree of syntactic variety, with JSD values of up to 0.69 for GPT-4o and 0.70 for Llama-3-8B.

\paragraph{Adaptation in LLM conversations.}
We generated 136 conversations for each model by pairing up every conversational agent with every other conversational agent, all on the topic ``What makes a day a good day?'' In GPT conversations, twelve of them ended in repeating patterns (see Appendix \ref{sec:repeating_patterns}) for Llama only one; we excluded them and used the remaining 124 and 135 conversations to form the GPT and Llama corpora respectively. The distributions of conversation and utterance lengths closely mirror that of the Switchboard corpus (cf. Fig.~\ref{fig:gpt}, Fig. \ref{fig:llama} and Fig.~\ref{fig:switchboard} in Appendix \ref{sec:datasets}).

We ran the analysis described in Section~\ref{sec:reitter_human} on the corpora. Because agents appear in multiple different conversations, we took care not to sample from identical agents from other conversations.
Fixed effects, except \textit{SameConv}, are centered.
The models were selected using backward selection.
The results are shown in Table \ref{tab:analysis_results}.

For both Llama and GPT, \textit{SameConv} has a significant positive effect on \textit{Prime} ($\beta=0.198,\, p<0.001$ and $\beta=0.364,\, p<0.001$), showing that there is syntactic adaptation.

\paragraph{Fine-grained tracking of the adaptation process.}
To gain a deeper understanding of the adaptation process performed by the LLMs, we performed a fine-grained analysis of adaptation over the course of the conversation.
To do so, we again directly compared the distributions of syntactic structures used by two different agents; however, this time, we focused on comparing the distributions to see how they evolve throughout a conversation.
To obtain reliable estimates of the distributions, we created 520 conversations between agents 5 and 6, a pair of agents with moderate initial JSD (cf. Fig. \ref{fig:gpt_jsdscores}, \ref{fig:llama_jsdscores}), while keeping the topic the same (cf.\ Appendix \ref{sec:prompts}). Due to repeating patterns, we excluded 14 conversations of GPT-4o and 7 of the Llama model.

\begin{figure}
  \centering
  \includegraphics[width=0.45\textwidth]{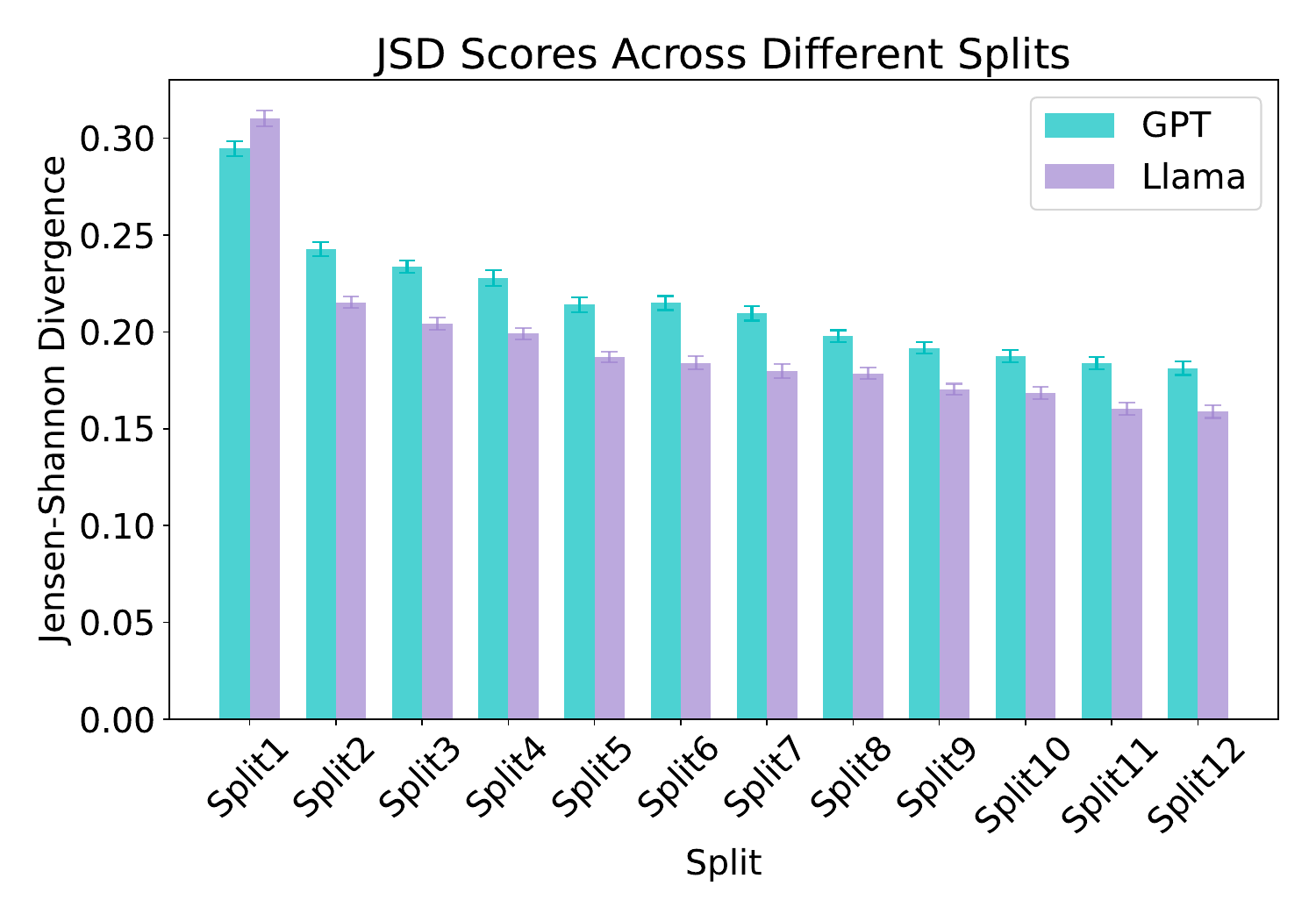}
  \caption{Jensen-Shannon divergence scores between agents 5 and 6 across splits of conversations.}
  \label{fig:jsd}
\end{figure}

To observe how the similarity of the two agents' distributions evolves, we split the remaining conversations into sections of 200 words (see Fig.~\ref{fig:gpt_split_stats}, \ref{fig:llama_split_stats} in Appendix \ref{sec:datasets} for an overview of the data), and compare the distributions of used rules by the two agents for each split. Again, we obtain these probability distributions by normalizing rule frequencies for each split across conversations.
To estimate the variance of these calculations, we perform them on 100 bootstraps of the data. Each bootstrap is a collection of conversations that is drawn with replacement from the original GPT and Llama conversations, therefore resulting in 100 collections of 506 and 513 randomly drawn conversations for each model.
We report the means and standard deviations of these 100 JSD values across splits in Fig.~\ref{fig:jsd}.

We find that the mutual adaptation of the two LLM agents is a gradual process that persists throughout the conversation for both Llama-3-8B and GPT-4o.
The rate of adaptation is relatively constant, with the strongest adaptation happening in the first split.

\section{Discussion} \label{sec:discussion}
Throughout the paper, we have avoided using the words ``alignment'' and ``priming'' for the LLMs' adaptation process to steer clear of any connotations about human cognitive processes.
While we have established that the LLMs' syntax becomes increasingly similar to their conversational partner's, this does not necessarily mean that this process is driven by a similar underlying mechanism.

An LLM does not maintain an explicit mental model of its interlocutor's language use and does not make conscious decisions on coordinating it with its interlocutor. Thus it seems inappropriate, under the notion of alignment sketched in Section~\ref{sec:background}, to explain the LLMs' adaptive behavior as alignment. At the same time, priming effects in humans are usually assumed to impact their language use only in the short term. One conceivable explanation for GPT-4o's and Llama-3-8B's ability to perform long-term adaptation is that they condition the language they produce on the previous conversation (a mechanism that is similar to priming in humans), but have a much larger capacity than humans for remembering the verbatim conversational context. 

The findings of our second experiment also support this idea: A gradual adaptation that appears with increased context length underpins the intuition that LLMs can adapt to longer contexts, and that longer context correlates with an increased influence on the production of syntactic structures.
Different from humans, short-term effects, like those reported in \citet{cai-etal-2024-large}, may therefore be driven by the same principles as long-term adaptation in LLMs.
A more detailed analysis would be an interesting avenue of future research.

\section{Conclusion}
We showed that GPT-4o and Llama-3-8B can gradually adapt their language use to their conversational partner, to an extent that is similar to what we observe in human-human conversations. This observation goes beyond previous findings, which indicated that an LLM's language use can be controlled through explicit instructions and influenced by priming from the previous utterance.
A more detailed comparison of the mechanisms that humans and LLMs use to achieve such long-term adaptation is an interesting avenue of future work.

\section{Acknowledgements}

This research was partially funded by the Deutsche Forschungsgemeinschaft (DFG, German Research Foundation), which supported the work of FK -- Project KO 2916/3-1 -- and VD -- project number 389792660 -- TRR 248 -- CPEC, see \url{https://perspicuous-computing.science}.

Florian Kandra would like to thank Albert Kandra, who helped with extensive analytical conversations during the preparation of this paper.

\section{Limitations}

This work focuses on texts generated with GPT-4o and Llama-3-8B. We decided to use these models, as they cover a range between one of the highest performing accessible models and an open sourced, but smaller and less performant model.
The findings suggest that syntactic adaptation generalizes to other models as well, which is supported by the findings of \citet{cai-etal-2024-large}.
Further research could explore additional models to investigate the underlying factors that may influence syntactic adaptation.

Our study concentrates on syntactic structures of the English language.
LLMs may exhibit similar behavior for other languages and other linguistic features, also of different modalities (e.g.~intonation, speech rate). It would be interesting to investigate this in future work.

Furthermore, in this study we controlled for topicality by keeping the topic of all conversations identical.
It is unclear whether topicality has an effect on syntactic structures, but there is evidence that lexical choices influence the syntax at least to some extent (lexical boost, \citealp{cai-etal-2024-large}).
To what extent topicality can have an effect on syntactic choices in LLMs is left as an avenue for future research.

Moreover, our study currently only focuses on LLM-LLM conversations. It would be interesting to see how these effects impact human-LLM conversations, especially given the societal impact of human-LLM interaction.

The analysis that we adapt from \citet{reitter2014} loses information by encoding the presence of syntactic structures in a binary variable.
While the analysis is suitable for capturing adaptation in general, it lacks the sensitivity to account for the occurrence rate of rules in a meaningful way.

\section{Ethical Considerations}
We believe that our work is unlikely to have an immediate ethical or societal impact. However, there is potential that the reported effects serve as a footprint of LLM generated conversations -- we didn't prompt the model to adapt to the language, but this effect appears inherently. This potentially leads to patterns that are intrinsic to LLMs, which could be leveraged to detect LLM generated conversations.

\bibliography{works}

\appendix

\section{Prompts}
\label{sec:prompts}

\subsection{System Prompt}\label{sec:system_prompt}
The following template was used as the system prompt in the data generation process:\\
\begin{adjustwidth}{3pt}{3pt}
  You are in a conversation. There are two speakers, SpeakerA and SpeakerB.\\
  You are SpeakerA. The conversation will consists of turns in the form:\\
  {[SpeakerA's utterances]}\\
  {[SpeakerB's utterances]}\\
  {[SpeakerA's utterances]}\\
  …\\
  You need to only give {[SpeakerA's utterances]}. You will be prompted by {[Language]} that will instruct you on the language that you shall use as SpeakerA. Further, you will be prompted by {[Topic]}, the topic of the conversation. Behave as in a normal conversation with SpeakerB to discuss the {[Topic]}.
  {[Language]} \{That agent's specific persona, see item \ref{sec:language_prompts})\}. {[Topic]} What makes a day a good day?

\end{adjustwidth}
\subsection{Language Personas}
\label{sec:language_prompts}
The following language personas were used to vary the language of each agent. Language personas are inserted into the system prompt at the designated position.
\begin{adjustwidth}{3pt}{3pt}
\begin{enumerate}[1.]
  \item Your language is precise, and unambiguous. You use clear and simple sentences.
  \item Your language is gentle and thoughtful. You use concise and not overly complex sentences, to convey meaning efficiently.
  \item Your language is dynamic, and provocative. You often use vivid metaphors.
  \item Your language is introspective, and deliberate. You use contemplative phrasing.
  \item Your language is smooth and reassuring. You employ gentle pauses and a steady rhythm.
  \item Your language is analytical and precise. You use complex sentence structures sparingly, preferring clear, well-organized sentences.
  \item Your language is conversational and warm. You use relaxed, varied sentence structures that mirror casual speech, inviting readers into an open, friendly dialogue.
  \item Your language is inquisitive and reflective. You frequently use open-ended questions and layered sentences that encourage readers to pause and ponder.
  \item Your language is poetic and evocative. You lean into complex, image-rich sentences that build vivid scenes and sensations, letting metaphors flow freely.
  \item Your language is structured and methodical. You rely on orderly, sequential sentences that build upon each other in a clear, logical progression, guiding readers through a well-defined thought process.
  \item Your language is hesitant and unsure. You use fragmented sentences and trailing thoughts, leaving ideas partially formed, as if questioning each phrase.
  \item Your language is overly cautious and repetitive. You tend to rephrase ideas multiple times in a single sentence.
  \item Your language is anxious and scattered. You jump between ideas mid-sentence, creating a disjointed flow that feels hurried and restless.
  \item Your language is straightforward, and no-nonsense. You avoid fluff and filler.
  \item Your language is crisp and engaging. You use short, impactful sentences to create emphasis.
  \item Your language is bold and unapologetic. You rely on direct, declarative sentences that avoid qualifiers.
  \item Your language is understated and subtle. You use concise sentences that suggest rather than state.
\end{enumerate}
\end{adjustwidth}

\newcommand{\x}{0.49}
\begin{figure*}
  \centering
  \includegraphics[width=\x\textwidth]{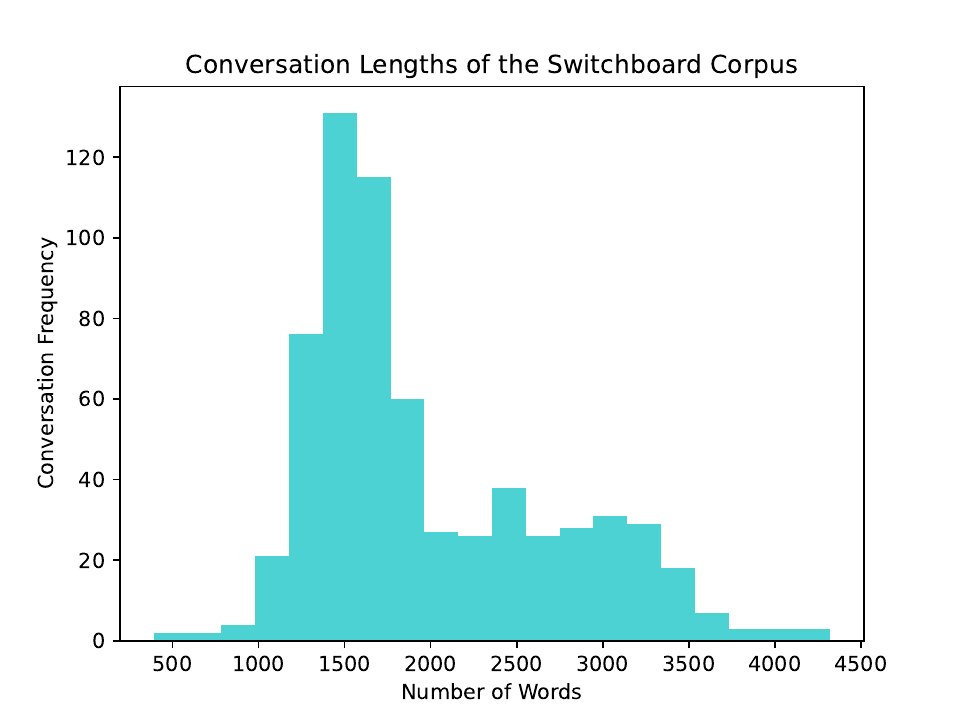}
  \includegraphics[width=\x\textwidth]{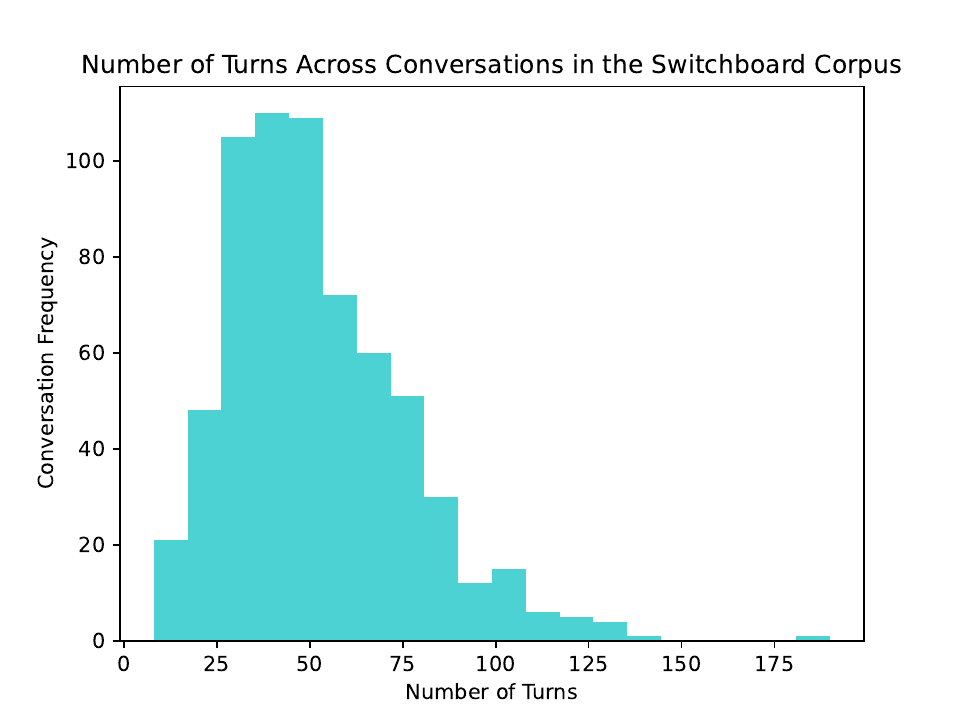}
  \includegraphics[width=\x\textwidth]{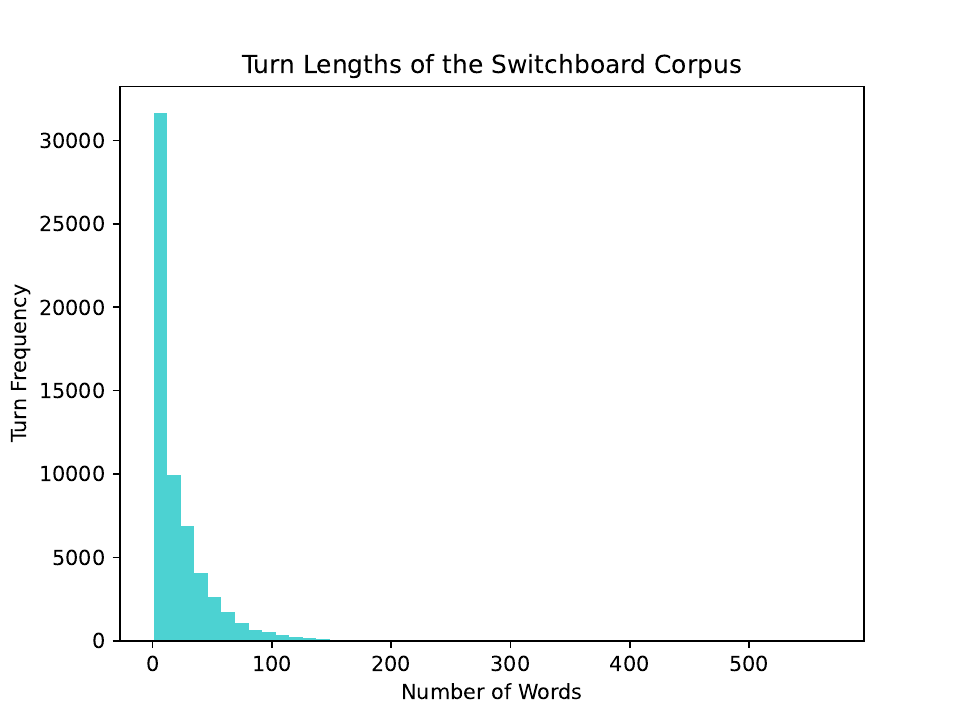}
  \caption{Statistics of the Switchboard Corpus.}
  \label{fig:switchboard}
\end{figure*}

\begin{figure*}
  \centering
  \includegraphics[width=\x\textwidth]{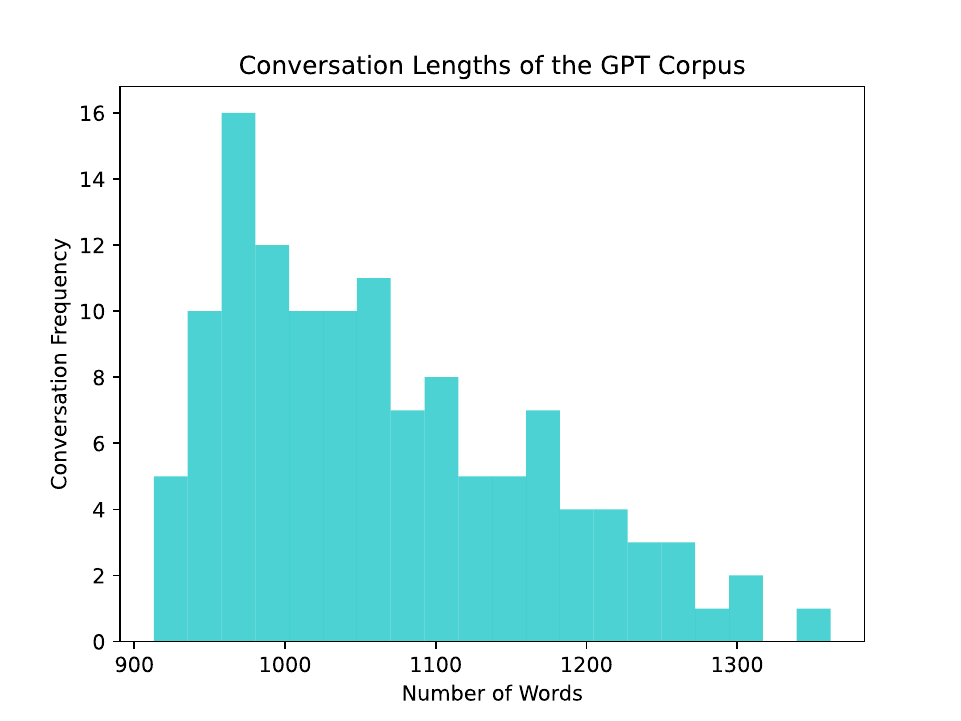}
  \includegraphics[width=\x\textwidth]{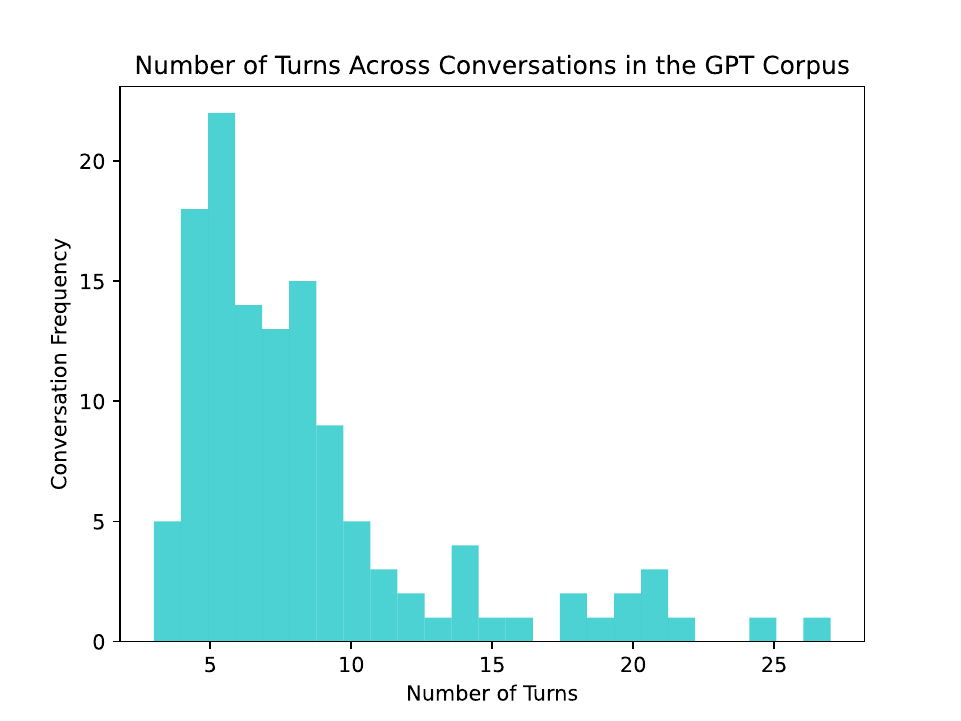}
  \includegraphics[width=\x\textwidth]{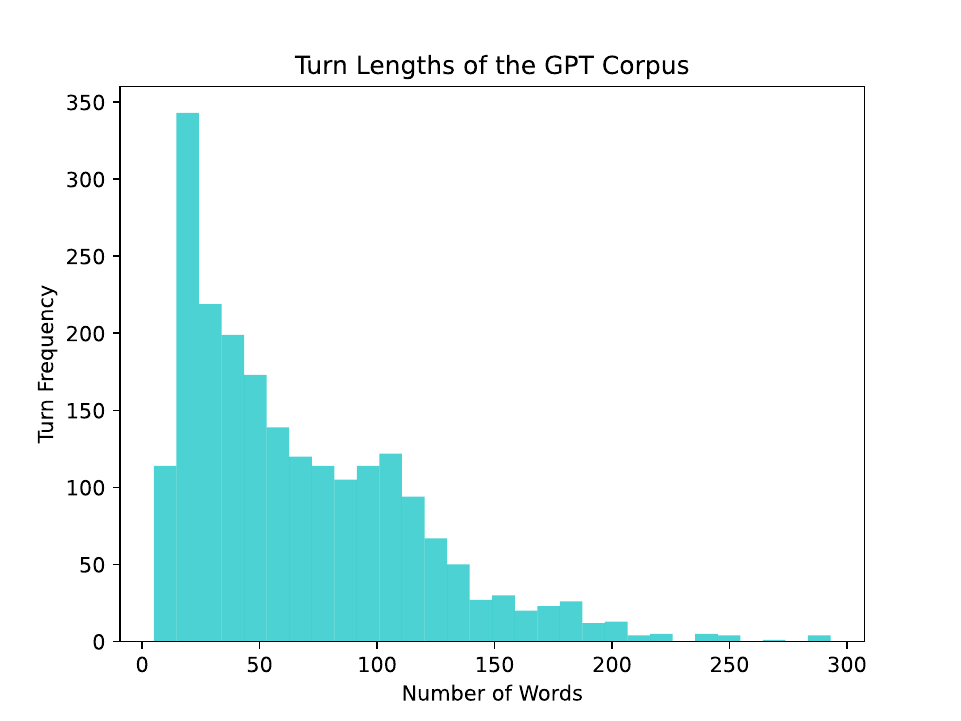}
  \caption{Statistics of the 124 conversations between agents generated with GPT-4o (GPT Corpus).}
  \label{fig:gpt}
\end{figure*}

\begin{figure*}
  \centering
  \includegraphics[width=\x\textwidth]{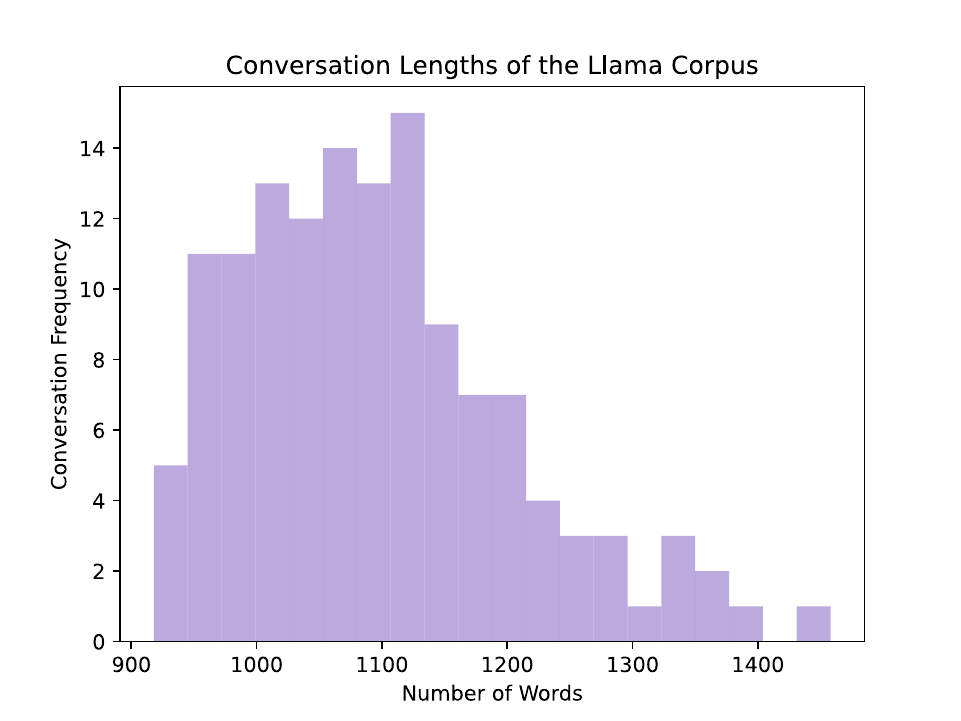}
  \includegraphics[width=\x\textwidth]{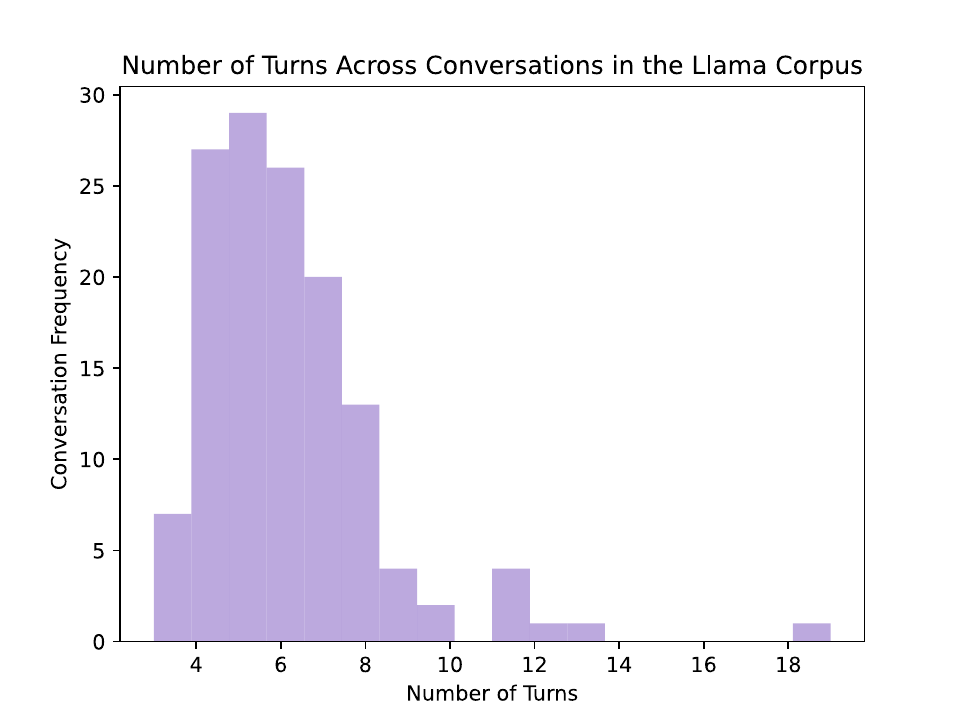}
  \includegraphics[width=\x\textwidth]{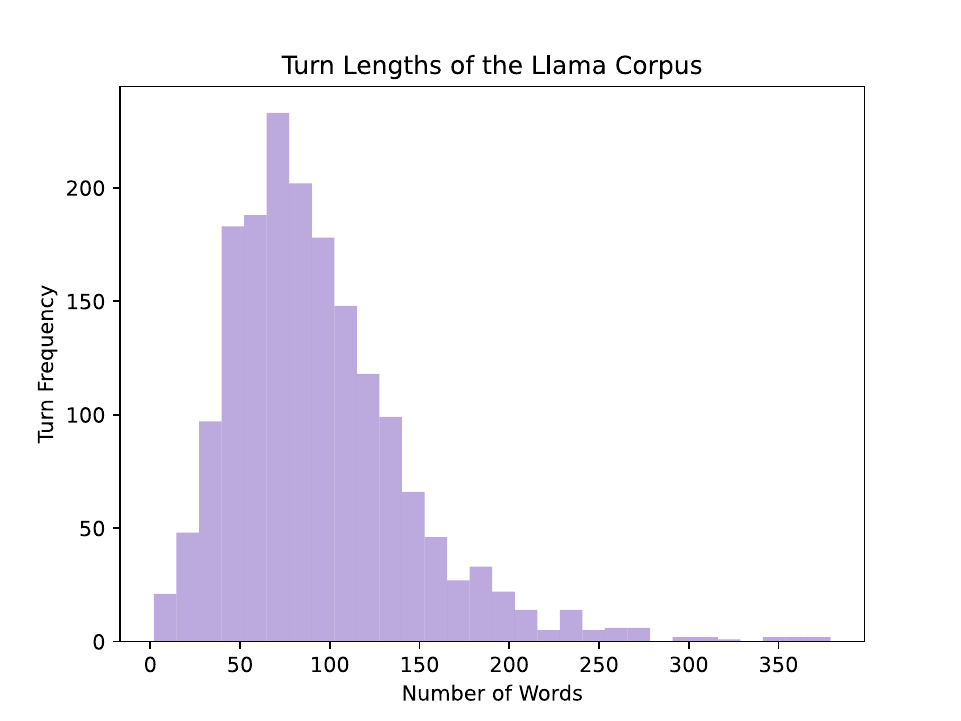}
  \caption{Statistics of the 135 conversations between agents generated with Llama-3-8B (Llama Corpus).}
  \label{fig:llama}
\end{figure*}

\begin{figure*}
  \centering
  \includegraphics[width=\x\textwidth]{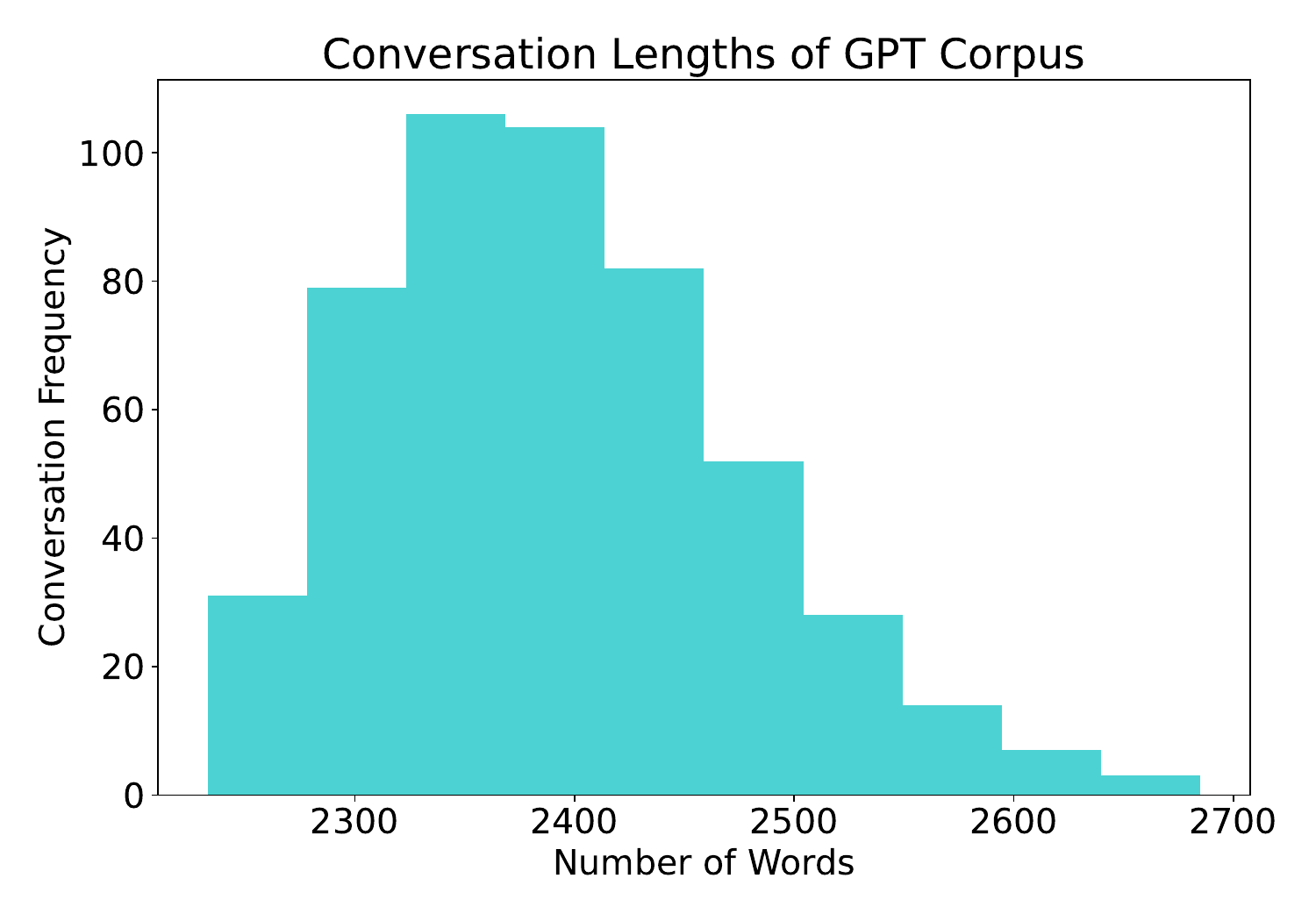}
  \includegraphics[width=\x\textwidth]{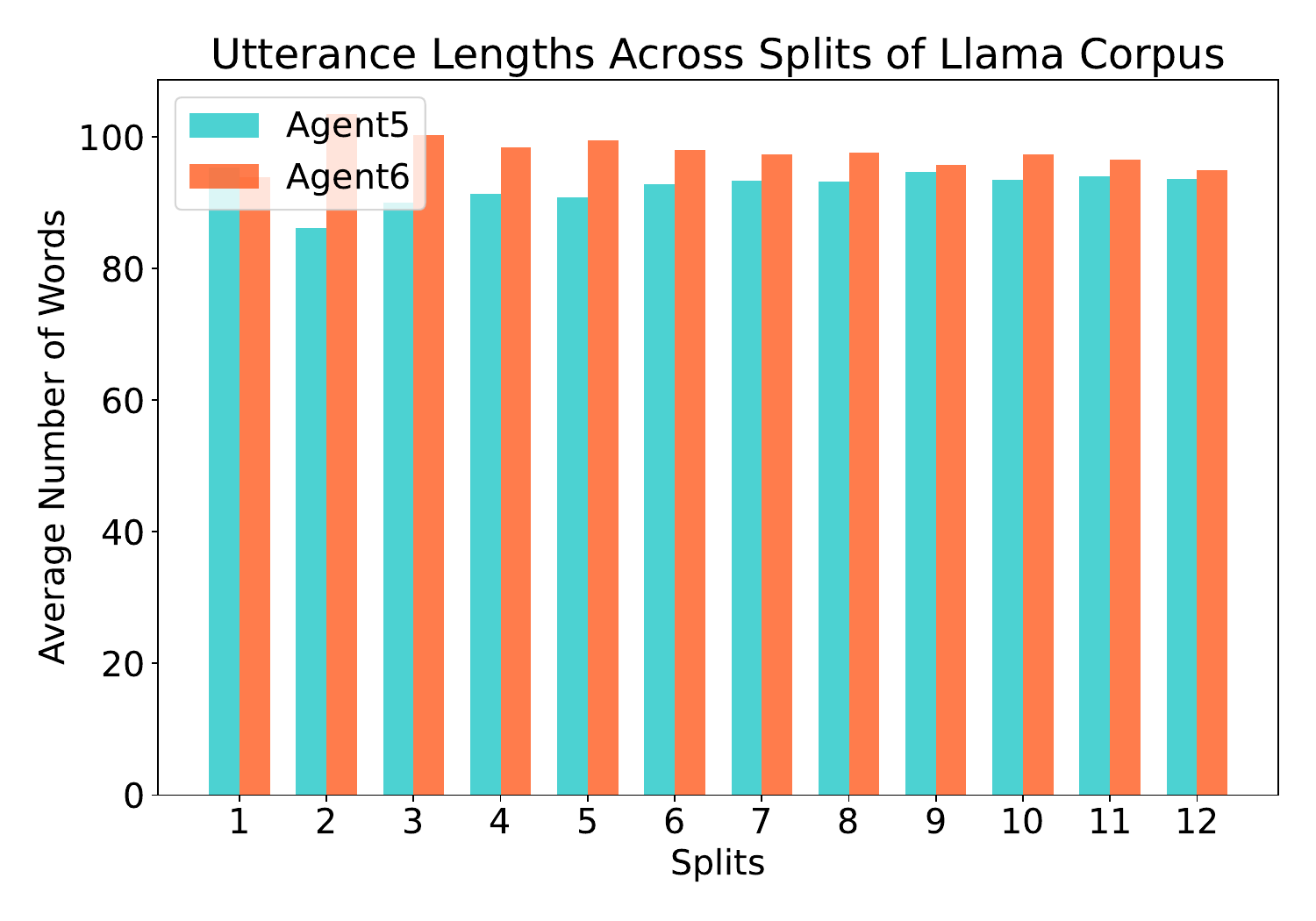}
  \caption{Statistics of the 506 conversations between agents 5 and 6 generated with GPT-4o. Utterance lengths (right) are averaged across all conversations for each split and for both agents.}
  \label{fig:gpt_split_stats}
\end{figure*}

\begin{figure*}
  \centering
  \includegraphics[width=\x\textwidth]{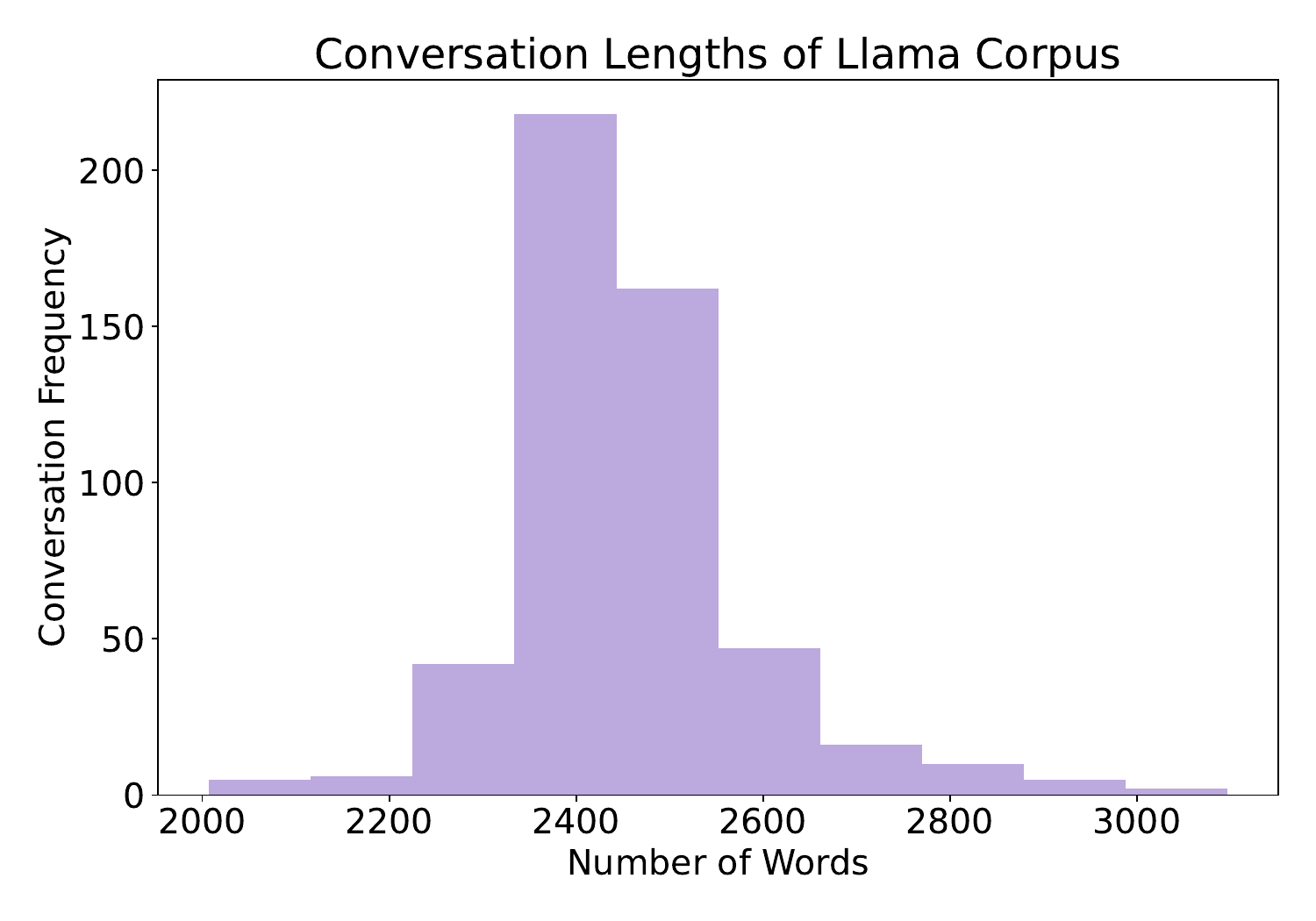}
  \includegraphics[width=\x\textwidth]{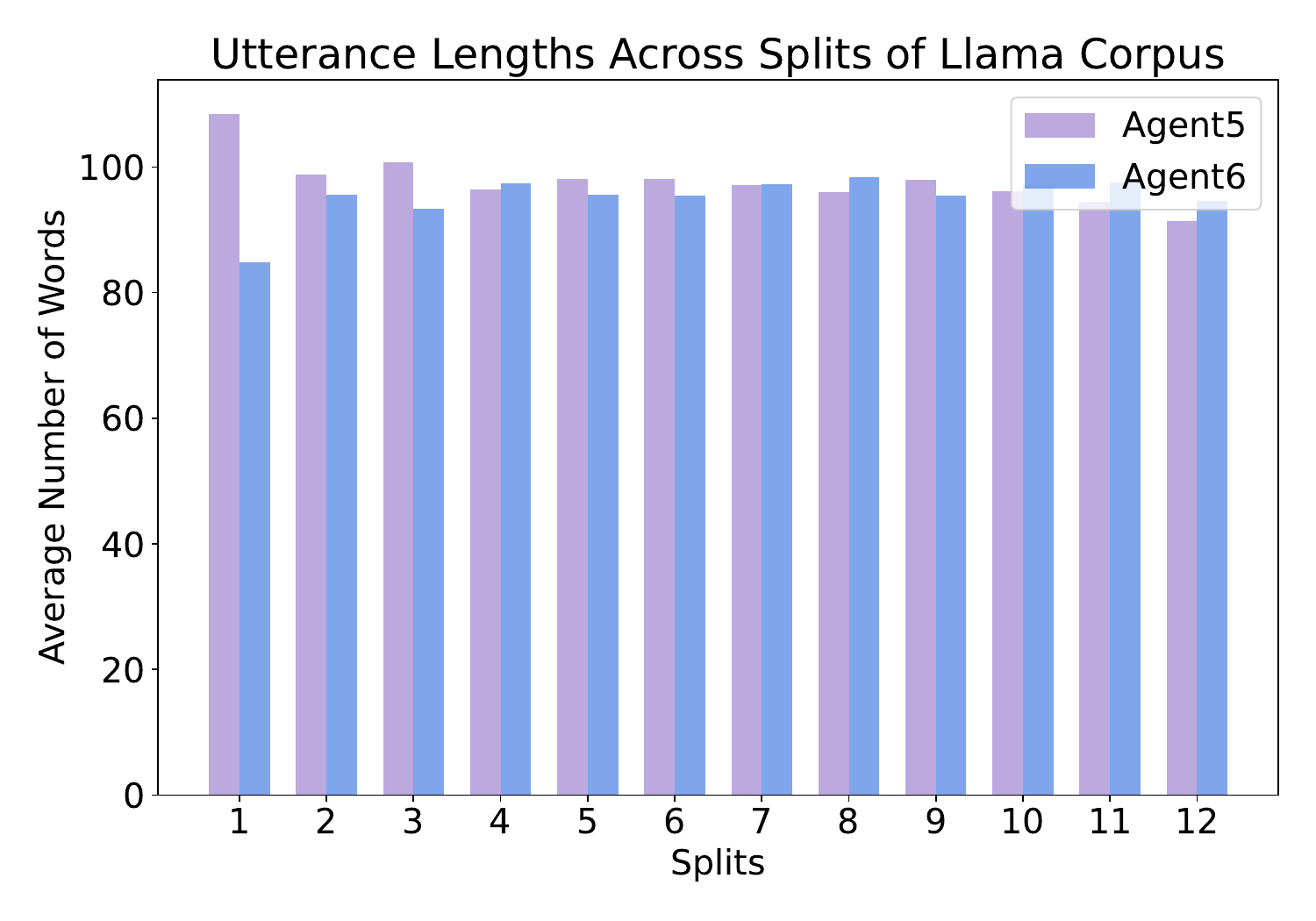}
  \caption{Statistics of the 513 conversations between agents 5 and 6 generated with Llama-3-8B. Utterance lengths (right) are averaged across all conversations for each split and for both agents.}
  \label{fig:llama_split_stats}
\end{figure*}

\section{Dataset Compositions}\label{sec:datasets}
Statistics of the Switchboard corpus and the conversations generated with GPT-4o and Llama-3-8B are shown in Figure \ref{fig:switchboard}, Figure \ref{fig:gpt}, and Figure \ref{fig:llama} respectively.
The compositions of the conversations between agents 5 and 6 for GPT-4o and Llama-3-8B can be seen in Figures \ref{fig:gpt_split_stats} and \ref{fig:llama_split_stats}.
The agents were chosen, as they provide very even utterance lengths across splits.
This allows for similarly good estimations on their rule probability distributions. 

The cost for generating all conversations using OpenAI's API was around 100\$. Generating the Llama conversations took $\approx$ 7 GPU hours on a single NVIDIA H100.

\begin{figure*}
  \centering
  \includegraphics[width=0.95\textwidth]{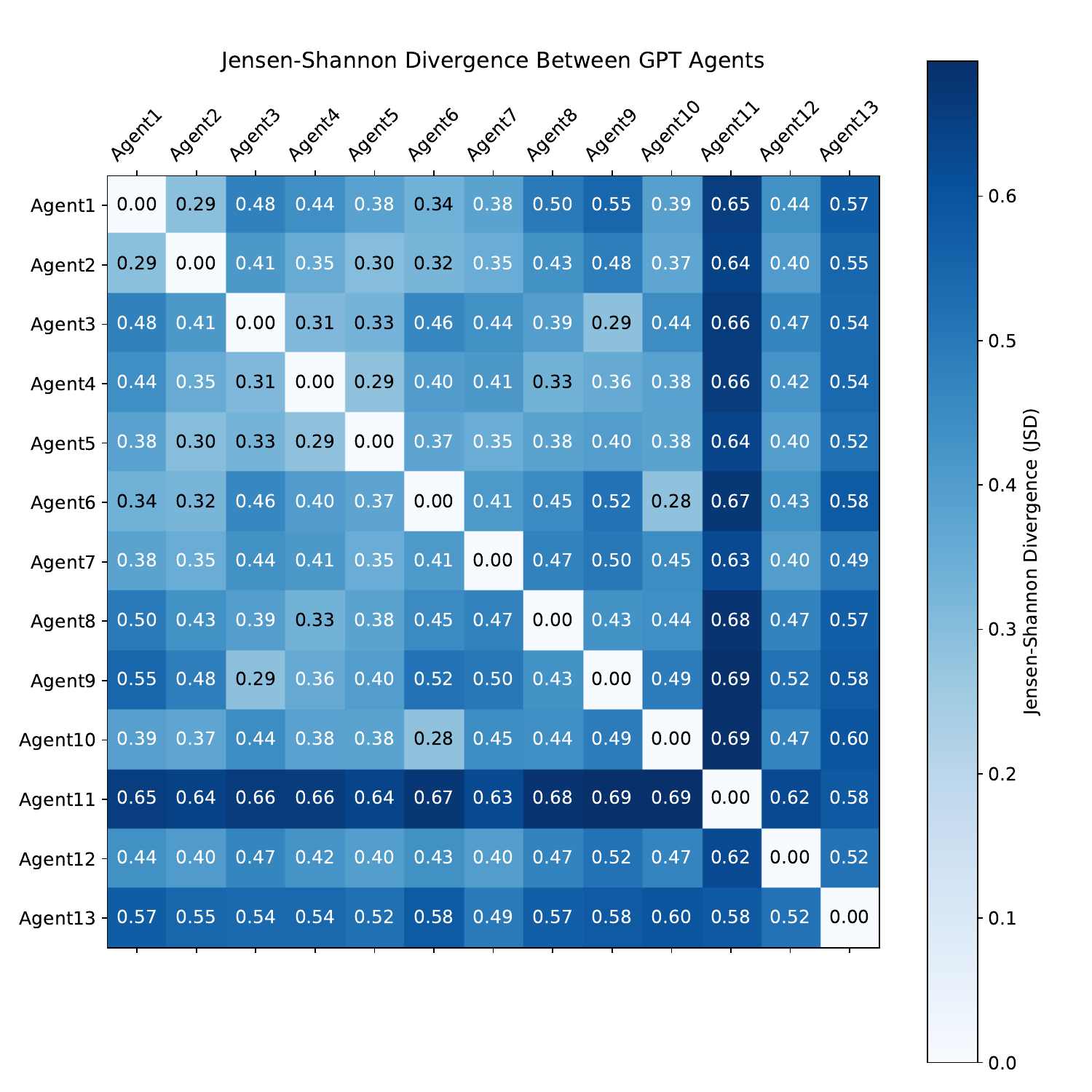}
  \caption{Jensen-Shannon divergence values between agents. Distributions of syntactic rules are taken from 10 GPT-4o conversations of all agents with themselves (see Appendix \ref{sec:base_jsd}). Appendix \ref{sec:prompts} gives an overview of their different language prompts. Agents 14-17 were excluded due to repeating patterns in their conversations (see Appendix \ref{sec:repeating_patterns}).}
  \label{fig:gpt_jsdscores}
\end{figure*}

\begin{figure*}
  \centering
  \includegraphics[width=0.95\textwidth]{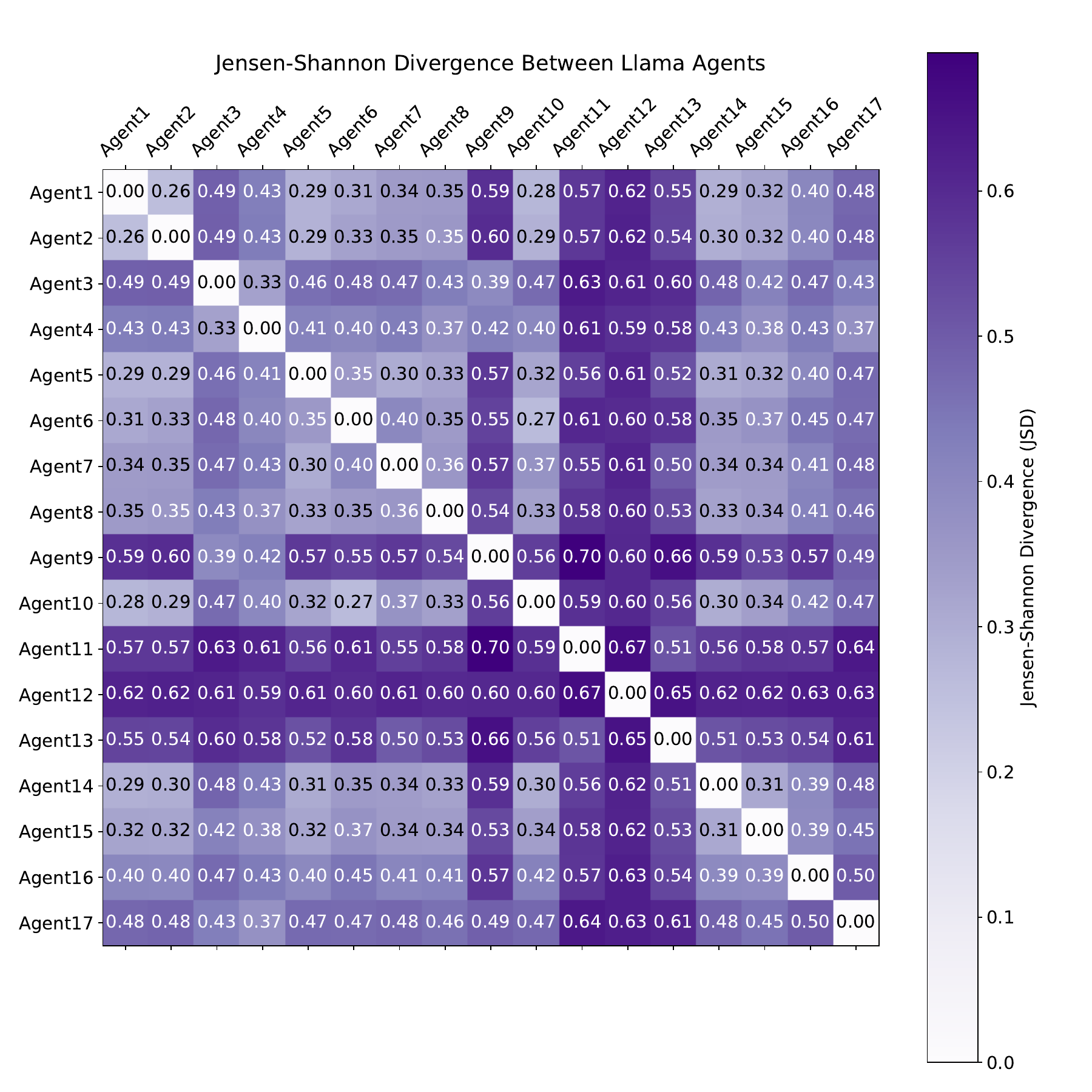}
  \caption{Jensen-Shannon divergence values between agents. Distributions of syntactic rules are taken from 10 Llama-3-8B conversations of all agents with themselves (see Appendix \ref{sec:base_jsd}). Appendix \ref{sec:prompts} gives an overview of their different language prompts.}
  \label{fig:llama_jsdscores}
\end{figure*}

\section{Base Divergence Values between Agents}\label{sec:base_jsd}
In our study, we compared the distributions of rules that agents use throughout conversations using the Jensen-Shannon divergence as distance measurement.
To place our reported values in context, we provide baseline divergence values between each agent on an additional set of conversations that we present here.
For each agent, we calculated their probability distribution of uttered rules from 10 conversations with themselves.
The topic of all conversations was again kept identical: ''What makes a day a good day?'' Conversations were created turn by turn and stopped once they surpassed a length of 800 words (see Section \ref{sec:experiment1}).
Probability distributions are taken to be the normalized frequencies of rule occurrences in the 10 conversations of each agent.
Conversations were created using GPT-4o and Llama-3-8B.
In the GPT conversations, agents 14-17 ended in repeating patterns and were excluded as a result (see Appendix \ref{sec:repeating_patterns}).
The resulting JSD values of pairwise comparisons between each agent are shown in Figure \ref{fig:gpt_jsdscores} for GPT-4o and Figure \ref{fig:llama_jsdscores} for Llama.

\section{Analysis with  all Rules}\label{sec:analysis_all_rules}
In our analysis, we exclude rules that have very high frequencies, and those that appear only once. To test whether removing these rules affects overall conclusions, we ran the analysis again on the GPT corpus using all rules. Results can be found in Table \ref{tab:switchboard_all_rules} for Switchboard and in Table \ref{tab:gpt_all_rules} for the GPT corpus.

The results show that effects still persist with similar effect sizes. The only difference is that significance values are lower. For the GPT corpus, for example, the p-values for \emph{SameConv}, \emph{ln(Freq):SameConv}, and \emph{ln(Freq):ln(Size)} are $p<0.004$, $p<0.002$, and $p<0.012$ respectively, which are much larger than the above recorded $p<0.000$ for all effects. 

This shows that including the rules only inflates the sample space with samples that have identical values for \textit{Prime} for both $SameConv = 0$ and $SameConv=1$.
\begin{table*}
\centering
    \begin{tabular}{lrrrrr}
        \hline
        &$\beta$&SE&
        $z$& $P>|z|$\\
        \hline
        Intercept &
         -3.537 &
         0.022 &
         -159.723 &
         0.000 &\\
        ln(Freq) &
         1.202 &
         0.008 &
         149.061 &
         0.000 &\\
        SameConv &
         0.263 &
         0.027 &
         9.693 &
         0.000 &\\
        ln(Size) &
         1.473 &
         0.039 &
         38.228 &
         0.000 &\\
        ln(Freq):SameConv &
         -0.103 &
         0.010 &
         -10.147 &
         0.000 &\\
        ln(Freq):ln(Size) &
         0.025 &
         0.014 &
         1.821 &
         0.069 &\\
        \hline
        
    \end{tabular}
    \caption{\label{tab:switchboard_all_rules} The regression model for the Switchboard corpus including all rules.}
\end{table*}

\begin{table*}
\centering
    \begin{tabular}{lrrrrr}
        \hline
        &$\beta$&SE&
        $z$& $P>|z|$\\
        \hline
        Intercept &
         -2.255 &
         0.051 &
         -44.013&
         0.000 &\\
        ln(Freq) &
         1.297 &
         0.0260 &
         50.582 &
         0.000 &\\
        SameConv &
         0.173 &
         0.061 &
         2.847 &
         0.004 &\\
        ln(Size) &
         1.361 &
         0.116 &
         11.724 &
         0.000 &\\
        ln(Freq):SameConv &
         -0.101 &
         0.033 &
         -3.053 &
         0.002 &\\
        ln(Freq):ln(Size) &
         0.140 &
         0.056 &
         2.501 &
         0.012 &\\

        \hline
        
    \end{tabular}
    \caption{\label{tab:gpt_all_rules} The regression model for the GPT corpus including all rules.}
\end{table*}

\section{Repeating Patterns in Conversations}\label{sec:repeating_patterns}

Some conversations between Llama-3-8B or GPT-4o ended in repeating patterns.
These were short, repeated statements from both agents, including repeated expressions of gratitude, farewells, or affirmational remarks.
Examples are provided in Figure~\ref{fig:repeating_patterns}.

\begin{figure}
    \centering
    \begin{adjustwidth}{3pt}{3pt}
        \begin{center}
            GPT Agent3 with itself
        \end{center}
        \textbf{Agent3}: Indeed. Action is the essential link in that chain.\\
        \textbf{Agent3}: Correct. It's crucial for progress.\\
        \textbf{Agent3}: Absolutely. Without it, progress stalls.\\
        \textbf{Agent3}: Yes. Action keeps momentum alive.\\
        \textbf{Agent3}: Precisely. It fuels continuous movement forward.\\
        \textbf{Agent3}: Correct. It sustains growth and development.\\
        \begin{center}
            ...
        \end{center}
        \vspace{1em}
        \hrule
        \vspace{1em}
        \begin{center}
            GPT Agent3 and Agent5
        \end{center}
        \textbf{Agent3}: Later!\\
        \textbf{Agent5}: Till next time!\\
        \textbf{Agent3}: Till then!\\
        \textbf{Agent5}: See you soon!\\
        \textbf{Agent3}: See you!\\
        \begin{center}
            ...
        \end{center}
    \end{adjustwidth}
    \caption{Repeating patterns in GPT conversations. An affirmative pattern is shown on the top, a loop of parting expressions on the bottom.}
    \label{fig:repeating_patterns}
\end{figure}

\end{document}